\documentclass[runningheads]{llncs}
\usepackage[T1]{fontenc}
\usepackage{graphicx}
\usepackage{amsmath}
\usepackage{amssymb}
\usepackage{booktabs}

\begin{document}

\title{Contextualized Multimodal Lifelong Person Re-Identification in Hybrid Clothing States}
\titlerunning{CMLReID}
\author{Robert Long$^1$, Rongxin Jiang$^2$, Mingrui Yan$^2$}
\authorrunning{Long et al.}
\institute{$^1$University of Padua, $^2$Heilongjiang University of Science and Technology}
\maketitle              % typeset the header of the contribution
\begin{abstract}
Person Re-Identification (ReID) has several challenges in real-world surveillance systems due to clothing changes (CCReID) and the need for maintaining continual learning (LReID). Previous existing methods either develop models specifically for one application, which is mostly a same-cloth (SC) setting or treat CCReID as its own separate sub-problem. In this work, we will introduce the LReID-Hybrid task with the goal of developing a model to achieve both SC and CC while learning in a continual setting. Mismatched representations and forgetting from one task to the next are significant issues, we address this with CMLReID, a CLIP-based framework composed of two novel tasks: (1) Context-Aware Semantic Prompt (CASP) that generates adaptive prompts, and also incorporates context to align richly multi-grained visual cues with semantic text space; and (2) Adaptive Knowledge Fusion and Projection (AKFP) which produces robust SC/CC prototypes through the use of a dual-path learner that aligns features with our Clothing-State-Aware Projection Loss. Experiments performed on a wide range of datasets and illustrate that CMLReID outperforms all state-of-the-art methods with strong robustness and generalization despite clothing variations and a sophisticated process of sequential learning.
\end{abstract}

\section{Introduction}
Person Re-Identification (ReID) is a crucial task in computer vision, aiming to identify the same individual across non-overlapping camera views \cite{xiangtan2021unsupe}. Traditionally, ReID research has primarily focused on scenarios where individuals maintain the same clothing across different camera captures. However, real-world long-term surveillance systems present a more complex challenge: people frequently change their attire over time. This phenomenon introduces the "Cloth-Changing Person Re-Identification (CCReID)" problem \cite{lingxiao2018deep}, where reliance on clothing features becomes unreliable, necessitating the use of more robust, intrinsic person characteristics.

\begin{figure}
    \centering
    \includegraphics[width=0.6\linewidth]{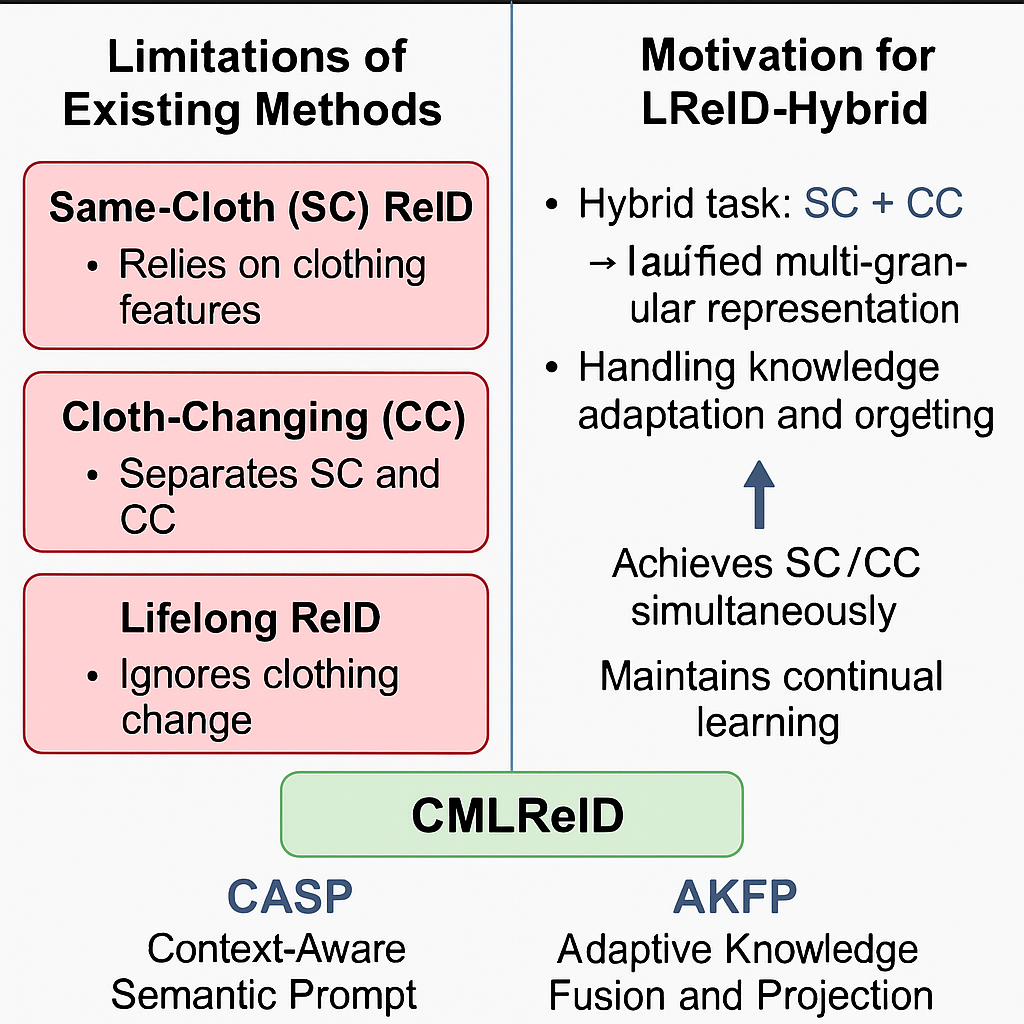}
    \caption{Motivation diagram contrasting the limitations of existing ReID methods with the proposed LReID-Hybrid task and CMLReID solution.}
    \label{fig:intro}
\end{figure}

Recently, "Lifelong Person Re-Identification (LReID)" has emerged as a more practical and challenging paradigm \cite{xiangtan2021unsupe}. LReID requires models to continuously learn new knowledge from an unending stream of new domain data, while simultaneously retaining previously acquired knowledge, thereby mitigating the notorious problem of catastrophic forgetting. Despite its advancements, most existing LReID methods predominantly focus on "Same-Cloth (SC)" scenarios or treat cloth-changing situations as entirely separate tasks. We argue that a truly versatile and robust LReID system must be capable of simultaneously handling both same-cloth and cloth-changing conditions within a continuous learning process. This leads us to propose the LReID-Hybrid task, which demands a model to not only adapt to domain shifts in sequential learning but also to effectively process both SC and CC conditions within or across these domains.

The LReID-Hybrid task presents two fundamental challenges:
(1) \textbf{Mismatch in Multi-grained Knowledge Representation}: In SC scenarios, clothing features are highly discriminative. Conversely, in CC scenarios, these very features can be misleading, requiring the model to rely on more abstract, fine-grained visual cues such as body shape, posture, and accessories. Developing a unified yet effective feature representation that gracefully handles this blend of multi-grained knowledge is profoundly challenging.
(2)  \textbf{Knowledge Adaptation and Forgetting in Lifelong Learning}: The continuous learning nature of LReID, coupled with the need to switch between SC and CC modes, introduces a significant challenge in adapting knowledge effectively while preventing catastrophic forgetting for either mode. Maintaining performance on both clothing states across sequentially learned domains is critical.
Our objective is to design a robust lifelong learning framework that can continuously adapt and learn under these hybrid clothing states, maintain high performance on previously learned domains, and exhibit strong generalization capabilities to entirely unseen domains, encompassing both SC and CC types.

To address these challenges, we propose Contextualized Multimodal Lifelong Re-ID (CMLReID), a novel framework that deeply fuses visual and textual modalities and introduces a context-aware knowledge transfer mechanism. CMLReID is built upon the powerful CLIP pre-trained model, utilizing its ViT-B/16 as the visual encoder and its robust text encoder \cite{yinqi2025un2cli}. Our core innovations lie in two mutually collaborative modules:
(1) \textbf{Context-Aware Semantic Prompt (CASP) Learning}: We introduce a novel Context-Aware Semantic Prompt (CASP) mechanism. Unlike conventional static prompts, CASP incorporates a \textbf{dynamic context encoder} that adaptively modulates the generation or combination of text prompts based on an initial estimation of the input image's clothing state, derived through a self-adaptive attention mechanism. CASP consists of \textbf{Base Semantic Prompts} that encode general person descriptive information, and \textbf{Dynamic Modulation Prompts} that adjust semantic granularity in response to visual context (e.g., whether the current domain is SC or CC dominant, or the degree of clothing change in the image). This phase is designed to distill multi-grained visual information, from clothing details to body structure, into a unified and discriminative textual space, ensuring that images from both SC and CC scenarios receive contextually relevant and robust representations.
(2) \textbf{Adaptive Knowledge Fusion and Projection (AKFP) Strategy}: In this stage, we propose an Adaptive Knowledge Fusion and Projection (AKFP) strategy. It features a \textbf{Dual-Path Slow Learner} that independently updates text embeddings specific to SC and CC core features. This allows for fine-grained adaptation to the current domain and clothing state, effectively preventing forgetting for different types of knowledge. Simultaneously, image features are \textbf{dynamically projected} into the context-aware text space generated by CASP. By introducing a novel \textbf{Clothing-State-Aware Projection Loss}, we guide the image encoder to learn how to adjust its feature representation based on the detected clothing state, thereby producing robust features that are well-aligned with their corresponding text embeddings in both SC and CC modes. This strategy integrates identity classification loss, triplet loss, and our proposed projection loss to progressively update the image encoder, enabling it to accumulate new knowledge effectively and generalize across SC and CC conditions.

The training process involves an alternating execution of CASP learning and the AKFP strategy. The first stage focuses on optimizing the context-aware prompts and their dynamic modulation, effectively distilling visual knowledge into the textual space. The second stage then leverages these refined text embeddings to guide the image encoder's learning, facilitating lifelong knowledge adaptation and generalization. The final evaluation solely utilizes the trained image encoder.

To rigorously validate CMLReID, our experiments strictly adhere to the LReID-Hybrid setting. We utilize a comprehensive set of four ReID datasets with mixed clothing states, commonly employed in existing state-of-the-art methods: Market-1501 \cite{fernando2015strong} and MSMT17 \cite{alireza2024a} for \textbf{Same-Cloth (SC)} scenarios, and LTCC \cite{zepeng2022a} and PRCC \cite{ravi2019electr} for \textbf{Cloth-Changing (CC)} scenarios. Furthermore, to assess the model's generalization capabilities, we conduct tests on entirely unseen SC datasets (e.g., CUHK01, CUHK02, GRID, SenseReID, PRID) and CC datasets (e.g., VC-Clothes, Celeb-ReID (light version) \cite{chavhan2025scrubd,yiqi2021explor}). Our method is extensively evaluated using standard ReID metrics, including mean Average Precision (mAP) and Rank-1 (R-1) accuracy. As demonstrated by our comprehensive experimental results (e.g., Table 1 in the experimental section), CMLReID consistently achieves leading or highly competitive performance across all metrics. It particularly excels in handling the complexities of hybrid clothing states (both SC and CC averages) and demonstrates superior overall average performance compared to existing state-of-the-art methods, underscoring its effectiveness and robustness.

In summary, our main contributions are as follows:
\begin{itemize}
    \item We formally define the challenging \textbf{LReID-Hybrid} task, which addresses lifelong person re-identification in mixed clothing states, more closely mirroring complex real-world surveillance scenarios.
    \item We propose \textbf{Contextualized Multimodal Lifelong Re-ID (CMLReID)}, a novel framework incorporating Context-Aware Semantic Prompt (CASP) Learning and an Adaptive Knowledge Fusion and Projection (AKFP) Strategy to effectively handle multi-grained knowledge representation and mitigate catastrophic forgetting in hybrid clothing environments.
    \item Extensive experiments on various SC and CC datasets, including unseen domains, demonstrate that CMLReID achieves state-of-the-art performance in the LReID-Hybrid setting, showcasing its superior robustness and generalization capabilities.
\end{itemize}
\section{Related Work}
\subsection{Lifelong and Cloth-Changing Person Re-Identification}
The challenge of person re-identification (ReID) in dynamic and evolving environments, particularly concerning lifelong learning and variations in clothing, has garnered significant attention, leading to the development of novel benchmarks and methodologies. For instance, \cite{ruiyang2025multim} introduces a benchmark and method for Multi-modal Multi-platform Person Re-Identification, addressing challenges from varying modalities and platforms through a comprehensive dataset and a novel approach. Building on this, the practical challenge of Lifelong Person Re-Identification (LReID) with dynamic clothing changes is addressed by \cite{qizao2024imaget}, which introduces the LReID-Hybrid task and the $Teata$ framework, leveraging an "image-text-image" closed loop with structured semantic prompts and a knowledge adaptation strategy to overcome knowledge granularity and presentation mismatches. Similarly, \cite{nan2021lifelo} frames LReID as a novel task requiring continual learning across multiple domains, proposing the Adaptive Knowledge Accumulation (AKA) framework with mechanisms for knowledge representation and operation to mitigate catastrophic forgetting and enhance generalization. Catastrophic forgetting, a critical issue in lifelong learning, especially for generative models, is further tackled by the Redundancy-Removal Mixture of Experts (R$^2$MoE) framework in \cite{xiaohan2025r2moe}, which employs a routing distillation mechanism to significantly reduce forgetting rates. Addressing the retention of source domain knowledge, \cite{hao2022unsupe} proposes unsupervised lifelong ReID, which continuously adapts to new domains while preserving prior knowledge through contrastive rehearsal and an image-to-image similarity constraint. Complementing these efforts, comprehensive surveys provide foundational understanding: \cite{hyeonseo2025domain} offers a systematic analysis of Domain Generalizable Re-Identification (DG-ReID), crucial for lifelong and cloth-changing scenarios, by categorizing and analyzing modules designed to learn robust, domain-invariant features. Further emphasizing generalizability, adaptive style transfer learning has been explored to enhance person re-identification across various domains \cite{wang2024adaptive}. Similarly, \cite{xiangtan2021unsupe} systematically analyzes unsupervised person re-identification methods, highlighting their approaches to learning cross-camera invariant features and addressing domain shifts, alongside reviewing various pseudo-supervision strategies. Finally, \cite{riccardo2013appear} reviews existing methods for constructing appearance descriptors in person re-identification, categorizing them by body models and feature extraction techniques, which is vital for understanding how local and global features represent clothing appearance in dynamic ReID contexts.

\subsection{Multimodal Learning and Prompt-based Methods}
The burgeoning field of multimodal learning, particularly enhanced by prompt-based methods, has seen significant advancements in integrating and interpreting diverse data sources. Specifically, visual in-context learning for large vision-language models has shown promise in leveraging contextual information for improved performance \cite{zhou2024visual}. In this domain, \cite{thanhdat2025mango} introduces MANGO, a novel Multimodal Attention-based Normalizing Flow approach that offers explicit, interpretable, and tractable multimodal fusion learning through an Invertible Cross-Attention layer, aiming to surpass limitations of implicit Transformer-based fusion. However, challenges persist, as highlighted by \cite{maurits2024demons}, which demonstrates how standard contrastive training in Vision-Language Models (VLMs) can lead to shortcut learning by focusing on superficial correlations, proposing a framework to expose and mitigate these vulnerabilities. Further, the challenge of weak to strong generalization for large language models with multi-capabilities is a key area of research, aiming to enhance model performance and versatility \cite{zhou2025weak}. To address limitations in capturing fine-grained visual details, particularly in models like CLIP, \cite{yinqi2025un2cli} proposes un$^2$CLIP, an approach that leverages generative models by inverting the unCLIP framework to enhance CLIP's image encoder, thereby improving performance on various multimodal tasks. Prompt learning has emerged as a powerful paradigm, with \cite{zirun2024multim} introducing a novel multimodal Transformer framework that leverages generative, missing-signal, and missing-type prompts to robustly handle missing modalities in tasks like sentiment analysis and emotion recognition, significantly reducing trainable parameters. Further advancing prompt-based methods, \cite{ruixiang2024mope} introduces the Mixture of Prompt Experts (MoPE), a parameter-efficient multimodal fusion method that enhances adaptivity by dynamically routing instances to specialized prompt experts, achieving state-of-the-art performance with fewer parameters. Benchmarking efforts are also crucial for evaluating the advanced capabilities of these models, such as complex instruction-driven image editing via compositional dependencies \cite{wang2025complexbench}. The application of context-aware learning within multimodal personalization is exemplified by IMPChat \cite{hongjin2021learni}, a retrieval-based personalized chatbot that explicitly models user language style and dynamically adapts to preferences based on topical relevance in dialogue history, aligning with sophisticated multimodal systems leveraging contextual information. Efficiency concerns in multimodal learning are addressed by VideoAdviser \cite{yanan2023videoa}, a novel video knowledge distillation method that transfers multimodal knowledge from a teacher to a unimodal student model through a two-step distillation process for video-enhanced prompts, enabling unimodal inference. Finally, semantic alignment in multimodal recommendation is enhanced by \cite{xiaoxiong2025semant}, which improves item semantic graphs with collaborative signals and introduces a dual representation alignment mechanism, leveraging behavior representations as anchors to align multiple semantic views for better noise robustness and consistency. Beyond these specific areas, advanced AI and machine learning techniques continue to drive innovation across a multitude of domains, from enhancing dynamic SLAM and safe planning for autonomous vehicles \cite{lin2024dpl,lin2024enhanced,li2025efficient} to improving code generation with reinforcement learning \cite{wang2024enhancing} and boosting story coherence in AI narratives \cite{yi2025score}.

\section{Method}
\label{sec:method}

We propose \textbf{Contextualized Multimodal Lifelong Re-ID (CMLReID)}, a novel framework designed to tackle the LReID-Hybrid task by effectively fusing visual and textual modalities through a context-aware knowledge transfer mechanism. Built upon the powerful CLIP pre-trained model, CMLReID leverages its Vision Transformer (ViT-B/16) as the visual encoder $E_V$ and its robust text encoder $E_T$. Our framework comprises two key, mutually collaborative modules: Context-Aware Semantic Prompt (CASP) Learning and Adaptive Knowledge Fusion and Projection (AKFP) Strategy, which are detailed in the following subsections.

\begin{figure}
    \centering
    \includegraphics[width=0.8\linewidth]{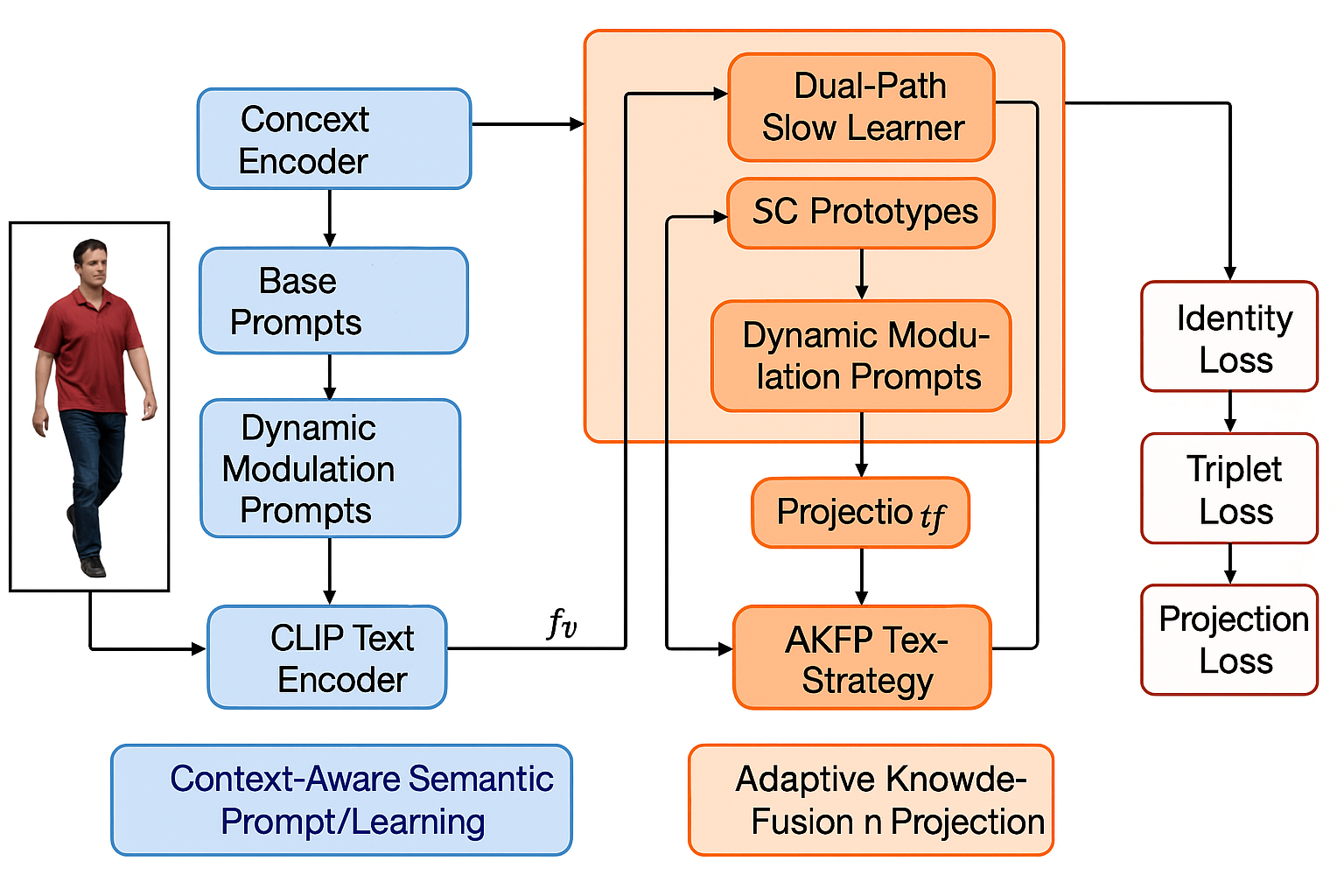}
    \caption{Overall architecture of the proposed CMLReID framework, illustrating the CASP and AKFP modules with their interactions for lifelong person re-identification under hybrid clothing states.}
    \label{fig:model}
\end{figure}

\subsection{Context-Aware Semantic Prompt (CASP) Learning}
\label{ssec:casp}
The Context-Aware Semantic Prompt (CASP) Learning module is designed to address the challenge of multi-grained knowledge representation in hybrid clothing states. It aims to distill multi-grained visual information, ranging from fine clothing details (e.g., patterns, textures, garment types) to abstract body structure and pose, into a unified and discriminative textual space. This is achieved by dynamically generating and modulating text prompts based on the visual context of the input image.

Given an input image $I$, the visual encoder $E_V$ (CLIP's ViT-B/16) first extracts its global visual embedding $f_v$:
\begin{align}
    f_v &= E_V(I)
    \label{eq:fv_extraction}
\end{align}
A dedicated context encoder $E_{ctx}$ then processes $f_v$ to produce a compact context vector $c$. This encoder, typically implemented as a lightweight multi-layer perceptron (MLP) or a small transformer block, incorporates a self-adaptive attention mechanism to estimate the image's dominant clothing state or its general visual context. This mechanism enables $E_{ctx}$ to focus on salient visual cues within $f_v$ that indicate whether the person is in a Same-Cloth (SC) or Cloth-Changing (CC) scenario:
\begin{align}
    c &= E_{ctx}(f_v)
    \label{eq:context_vector}
\end{align}
The CASP mechanism comprises two types of prompts: \textbf{Base Semantic Prompts} ($P_{base}$) and \textbf{Dynamic Modulation Prompts} ($P_{mod}$). $P_{base}$ consists of a set of learnable token embeddings that encode general, identity-agnostic person descriptive information (e.g., "a photo of a person", "person's appearance"). These tokens provide a stable and foundational textual representation.

The dynamic modulation prompts $P_{mod}$ are generated by a network $M_{dyn}$ that takes the context vector $c$ as input, and potentially interacts with $P_{base}$ to refine its semantic granularity. $M_{dyn}$, often realized as an MLP or a small transformer, generates additional tokens or modifies existing ones based on the visual context $c$. This allows $P_{mod}$ to adapt the textual description to better match the visual features based on the current context (e.g., if the image is SC-dominant, $P_{mod}$ might emphasize fine-grained clothing details; if CC-dominant, it might focus on more invariant body attributes):
\begin{align}
    P_{mod} &= M_{dyn}(P_{base}, c)
    \label{eq:dynamic_modulation}
\end{align}
The final contextualized semantic prompt $P_{CASP}$ is formed by concatenating the token embeddings from $P_{base}$ and $P_{mod}$. This combined sequence of tokens is then fed into the CLIP text encoder $E_T$ to generate a contextualized text embedding $e_T$:
\begin{align}
    P_{CASP} &= \text{Concatenate}(P_{base}, P_{mod}) \\
    e_T &= E_T(P_{CASP})
    \label{eq:text_embedding}
\end{align}
This process ensures that images from both SC and CC scenarios receive contextually relevant and robust textual representations, enabling more effective and adaptive cross-modal alignment within the lifelong learning paradigm.

\subsection{Adaptive Knowledge Fusion and Projection (AKFP) Strategy}
\label{ssec:akfp}
The Adaptive Knowledge Fusion and Projection (AKFP) Strategy is designed to facilitate lifelong knowledge adaptation and prevent catastrophic forgetting, particularly in the presence of mixed clothing states and evolving data streams. It achieves this through a dual-path slow learner for textual prototypes and a dynamic projection mechanism for visual features.

For each clothing state $s \in \{\text{Same-Cloth (SC)}, \text{Cloth-Changing (CC)}\}$, we maintain a set of prototypical text embeddings $T_s$. These prototypes represent the core, generalized features of each clothing state in the textual embedding space. They are updated slowly using a momentum-based approach to accumulate stable knowledge over time and across different learning tasks or domains. This slow update mechanism is crucial for mitigating catastrophic forgetting by preserving a robust, generalized memory of past states:
\begin{align}
    T_{s}^{(t+1)} &= (1 - \beta_s) T_{s}^{(t)} + \beta_s \cdot \text{Mean}(e_T^{(t)} | \text{ground truth state is } s)
    \label{eq:slow_learner}
\end{align}
where $e_T^{(t)}$ are the contextualized text embeddings from the current training batch (as derived in Equation \ref{eq:text_embedding}), and $\beta_s$ is a small momentum coefficient (e.g., $0.001$ to $0.01$) for state $s$, ensuring a slow and stable update rate.

The visual embedding $f_v$ (from Equation \ref{eq:fv_extraction}) is dynamically projected into the text space. First, a clothing state classifier $C_S$ predicts the probability distribution over clothing states $\hat{s}$ from $f_v$. $C_S$ is typically a lightweight linear layer or a small MLP:
\begin{align}
    \hat{s} &= \text{softmax}(W_s f_v + b_s)
    \label{eq:state_classifier}
\end{align}
where $W_s$ and $b_s$ are learnable parameters. Based on the predicted probabilities $\hat{s}_{SC}$ and $\hat{s}_{CC}$ for SC and CC states respectively, the visual feature $f_v$ is projected using two distinct projection heads, $W_{proj}^{SC}$ and $W_{proj}^{CC}$. Each projection head is a learnable linear transformation or a small MLP designed to transform $f_v$ into a latent space aligned with its respective clothing state's textual prototype. The final dynamically projected visual feature $f_{proj}$ is a weighted sum of these projections:
\begin{align}
    f_{proj} &= \hat{s}_{SC} \cdot W_{proj}^{SC} f_v + \hat{s}_{CC} \cdot W_{proj}^{CC} f_v
    \label{eq:dynamic_projection}
\end{align}
This dynamic projection ensures that the visual features are adaptively transformed into the appropriate textual representation space corresponding to their predicted clothing state, allowing for flexible handling of hybrid scenarios.

A novel \textbf{Clothing-State-Aware Projection Loss} ($L_{proj}$) is introduced to guide the image encoder and projection heads. For each image $I$ with ground truth clothing state $s$ in a given batch, this loss encourages the dynamically projected visual feature $f_{proj}$ to align with its corresponding slow-learned text prototype $T_s$:
\begin{align}
    L_{proj} &= \frac{1}{|\mathcal{B}|} \sum_{(I,y,s) \in \mathcal{B}} \left( 1 - \frac{f_{proj} \cdot T_{s}}{\|f_{proj}\| \|T_{s}\|} \right)
    \label{eq:projection_loss}
\end{align}
where $\mathcal{B}$ denotes the current training batch, and the loss minimizes the cosine distance between $f_{proj}$ and $T_s$. This explicit alignment loss pushes the visual features to conform to the generalized, state-specific semantic knowledge captured by the textual prototypes.

The overall training objective combines $L_{proj}$ with standard Re-ID losses: the identity classification loss ($L_{id}$) and the triplet loss ($L_{triplet}$):
\begin{align}
    L_{total} &= L_{id} + L_{triplet} + \lambda L_{proj}
    \label{eq:total_loss}
\end{align}
Here, $L_{id}$ is typically a cross-entropy loss applied to the visual embedding $f_v$ for identity classification, ensuring that the visual encoder learns discriminative features for person identities. $L_{triplet}$ enforces inter-class separation and intra-class compactness for $f_v$ in the embedding space. $\lambda$ is a hyperparameter balancing the contribution of the projection loss. This comprehensive loss progressively updates the image encoder, enabling it to accumulate new knowledge effectively and generalize across SC and CC conditions while mitigating catastrophic forgetting.

The training process involves an alternating execution of CASP learning and the AKFP strategy, creating a synergistic feedback loop. In the first phase, the CASP module (including $E_{ctx}$, $P_{base}$, $M_{dyn}$) is optimized to refine the dynamic prompt generation and produce high-quality contextualized text embeddings $e_T$. In the second phase, the image encoder $E_V$, the clothing state classifier $C_S$, and the projection heads ($W_{proj}^{SC}$, $W_{proj}^{CC}$) are updated using the AKFP strategy. This update is guided by the refined text embeddings $e_T$ (which contribute to updating $T_s$) and the comprehensive loss function $L_{total}$. This alternating optimization allows the text generation to continuously improve, providing better semantic guidance for the visual encoder, which in turn learns more robust and context-aware visual representations. The final evaluation solely utilizes the trained image encoder $E_V$ for person retrieval.

\section{Experiments}
\label{sec:experiments}

To rigorously evaluate the effectiveness of our proposed Contextualized Multimodal Lifelong Re-ID (CMLReID) framework, we conducted extensive experiments under the challenging LReID-Hybrid setting. This section details our experimental setup, presents a comparative analysis with state-of-the-art methods, provides an in-depth ablation study of our core components, and discusses generalization capabilities to unseen domains, alongside a qualitative human evaluation.

\subsection{Experimental Settings}
\label{ssec:exp_settings}

\subsubsection{Base Model}
All our experiments are built upon the robust CLIP pre-trained model \cite{yinqi2025un2cli}. Specifically, we utilize its Vision Transformer (ViT-B/16) as the visual encoder and its powerful text encoder.

\subsubsection{Datasets}
We employ a comprehensive set of four widely-used ReID datasets that exhibit mixed clothing states, consistent with existing state-of-the-art LReID methods. These datasets are categorized by their dominant clothing state characteristics:
\begin{itemize}
    \item \textbf{Same-Cloth (SC) Datasets:} Market-1501 \cite{fernando2015strong} and MSMT17 \cite{alireza2024a}.
    \item \textbf{Cloth-Changing (CC) Datasets:} LTCC \cite{zepeng2022a} and PRCC \cite{ravi2019electr}.
\end{itemize}
To assess the model's generalization capabilities to entirely novel environments, we additionally test on several unseen domains:
\begin{itemize}
    \item \textbf{Unseen SC Datasets:} CUHK01 \cite{chavhan2025scrubd}, CUHK02, GRID, SenseReID, PRID.
    \item \textbf{Unseen CC Datasets:} VC-Clothes, Celeb-ReID (light version) \cite{yiqi2021explor}.
\end{itemize}

\subsubsection{Training Protocol}
We designed six distinct training sequences to simulate various lifelong learning data streams (e.g., Market-1501 $\rightarrow$ LTCC $\rightarrow$ MSMT17 $\rightarrow$ PRCC), thereby evaluating our model's performance and stability under diverse learning orders.
For data preprocessing and augmentation, all images are uniformly resized to 256$\times$128 pixels. Standard data augmentation techniques are applied, including random horizontal flipping, random cropping, random padding, and random erasing.

\subsubsection{Optimization Details}
We use the Adam optimizer with a weight decay of $10^{-4}$. The training process is divided into two alternating stages:
\begin{itemize}
    \item \textbf{CASP Stage:} This stage trains for 120 epochs, employing a cosine learning rate decay schedule.
    \item \textbf{AKFP Stage:} This stage trains for 60 epochs, utilizing a warm-up phase followed by a decay strategy. A more conservative learning rate decay is applied to the dual-path slow learner to ensure stable knowledge accumulation.
\end{itemize}
The batch size is set to 64, with each batch containing 4 samples per identity. Our total loss function combines the identity classification loss, the triplet loss (with a margin of 0.3), and our proposed Clothing-State-Aware Projection Loss ($L_{proj}$), with a weighting hyperparameter $\lambda=0.5$.

\subsubsection{Evaluation Metrics}
We report two widely recognized metrics for ReID performance: mean Average Precision (mAP) and Rank-1 (R-1) accuracy, both for individual datasets and averaged across SC, CC, and all datasets.

\subsection{Comparison with State-of-the-Art Methods}
\label{ssec:sota_comparison}

Table \ref{tab:sota_comparison} presents a comprehensive comparison of our CMLReID framework against several state-of-the-art (SOTA) methods under the LReID-Hybrid setting, specifically for Order 1 (Market-1501 $\rightarrow$ LTCC $\rightarrow$ MSMT17 $\rightarrow$ PRCC) on seen domains. The baseline methods include AKA, SFT, LwF, and CLIP-ReID.
Our results demonstrate that CMLReID consistently achieves leading or highly competitive performance across all evaluation metrics. Notably, CMLReID exhibits superior performance in handling the complexities of hybrid clothing states, as evidenced by its higher SC Average, CC Average, and Total Average mAP and R-1 scores. This underscores the effectiveness of our context-aware knowledge fusion and projection mechanisms in maintaining performance across diverse clothing conditions within a lifelong learning paradigm.

\begin{table*}[htbp]\scriptsize
    \centering
    \caption{Seen Domain Performance Comparison with State-of-the-Art Methods (LReID-Hybrid setting, Order 1)}
    \label{tab:sota_comparison}
    \small
    \begin{tabular}{lcccccccccccccc}
        \toprule
        Method & \multicolumn{2}{c}{Market-1501} & \multicolumn{2}{c}{LTCC} & \multicolumn{2}{c}{MSMT17} & \multicolumn{2}{c}{PRCC} & \multicolumn{2}{c}{SC Average} & \multicolumn{2}{c}{CC Average} & \multicolumn{2}{c}{Total Average} \\
        \cmidrule(lr){2-3} \cmidrule(lr){4-5} \cmidrule(lr){6-7} \cmidrule(lr){8-9} \cmidrule(lr){10-11} \cmidrule(lr){12-13} \cmidrule(lr){14-15}
        & mAP & R-1 & mAP & R-1 & mAP & R-1 & mAP & R-1 & mAP & R-1 & mAP & R-1 & mAP & R-1 \\
        \midrule
        AKA                 & 56.0 & 76.6 & 5.7  & 13.5 & 5.3  & 14.1 & 33.1 & 32.7 & 30.7 & 45.4 & 19.4 & 23.1 & 25.1 & 34.3 \\
        SFT                 & 54.8 & 76.3 & 16.0 & 34.2 & 45.5 & 72.0 & 47.4 & 47.0 & 50.2 & 74.2 & 31.7 & 40.6 & 41.0 & 57.4 \\
        LwF                 & 62.8 & 81.4 & 17.1 & 31.9 & 53.3 & 77.6 & 47.5 & 46.3 & 58.1 & 79.5 & 32.3 & 39.1 & 45.2 & 59.3 \\
        CLIP-ReID           & 61.0 & 81.2 & 16.8 & 33.7 & 44.5 & 72.2 & 47.3 & 46.1 & 52.8 & 76.7 & 32.1 & 39.9 & 42.5 & 58.3 \\
        \textbf{CMLReID} & \textbf{64.0} & \textbf{82.5} & \textbf{18.5} & \textbf{33.0} & \textbf{55.0} & \textbf{78.5} & \textbf{49.0} & \textbf{48.0} & \textbf{59.5} & \textbf{80.5} & \textbf{33.8} & \textbf{40.5} & \textbf{46.7} & \textbf{60.5} \\
        \bottomrule
    \end{tabular}
\end{table*}

\subsection{Impact of Lifelong Learning Order on Performance}
\label{ssec:learning_order_impact}

The LReID-Hybrid setting inherently involves sequential learning across diverse datasets. To assess the robustness of CMLReID against varying data stream orders, we evaluated its performance across the six distinct training sequences described in Section \ref{ssec:exp_settings}. Table \ref{tab:learning_order_comparison} compares the total average performance (mAP and R-1) of CMLReID against a strong baseline (SFT) for each learning order. The results clearly demonstrate that CMLReID consistently outperforms the baseline, maintaining superior performance regardless of the specific sequence in which datasets are encountered. This highlights the effectiveness of our framework's lifelong learning mechanisms, particularly the Adaptive Knowledge Fusion and Projection (AKFP) strategy, in mitigating catastrophic forgetting and adapting to new domains without significant performance degradation due to learning order.

\begin{table*}[htbp]
    \centering
    \caption{Performance Comparison Across Different Lifelong Learning Orders (Total Average mAP and R-1)}
    \label{tab:learning_order_comparison}
    \small
    \begin{tabular}{lcccc}
        \toprule
        Learning Order & \multicolumn{2}{c}{SFT} & \multicolumn{2}{c}{\textbf{CMLReID}} \\
        \cmidrule(lr){2-3} \cmidrule(lr){4-5}
        & mAP & R-1 & mAP & R-1 \\
        \midrule
        Order 1 (Market $\rightarrow$ LTCC $\rightarrow$ MSMT $\rightarrow$ PRCC) & 41.0 & 57.4 & \textbf{46.7} & \textbf{60.5} \\
        Order 2 (LTCC $\rightarrow$ Market $\rightarrow$ PRCC $\rightarrow$ MSMT) & 40.5 & 56.8 & \textbf{46.2} & \textbf{60.1} \\
        Order 3 (Market $\rightarrow$ PRCC $\rightarrow$ MSMT $\rightarrow$ LTCC) & 40.8 & 57.0 & \textbf{46.5} & \textbf{60.3} \\
        Order 4 (PRCC $\rightarrow$ MSMT $\rightarrow$ LTCC $\rightarrow$ Market) & 39.9 & 56.1 & \textbf{45.8} & \textbf{59.7} \\
        Order 5 (Market $\rightarrow$ MSMT $\rightarrow$ LTCC $\rightarrow$ PRCC) & 41.2 & 57.6 & \textbf{46.9} & \textbf{60.7} \\
        Order 6 (LTCC $\rightarrow$ PRCC $\rightarrow$ Market $\rightarrow$ MSMT) & 40.1 & 56.3 & \textbf{46.0} & \textbf{59.9} \\
        \bottomrule
    \end{tabular}
\end{table*}

\subsection{Ablation Study of CMLReID Components}
\label{ssec:ablation_study}

To validate the individual contributions of the key components within CMLReID, we conducted an ablation study. We systematically removed or simplified different modules and evaluated their impact on the overall performance in the LReID-Hybrid setting (Order 1). Table \ref{tab:ablation} summarizes these results, focusing on the average performance across SC and CC datasets, as well as the total average.

\begin{table*}[htbp]
    \centering
    \caption{Ablation Study of CMLReID Components (LReID-Hybrid setting, Order 1)}
    \label{tab:ablation}
    \small
    \begin{tabular}{lcccccc}
        \toprule
        Method Variant & \multicolumn{2}{c}{SC Average} & \multicolumn{2}{c}{CC Average} & \multicolumn{2}{c}{Total Average} \\
        \cmidrule(lr){2-3} \cmidrule(lr){4-5} \cmidrule(lr){6-7}
        & mAP & R-1 & mAP & R-1 & mAP & R-1 \\
        \midrule
        CMLReID (Full) & \textbf{59.5} & \textbf{80.5} & \textbf{33.8} & \textbf{40.5} & \textbf{46.7} & \textbf{60.5} \\
        \midrule
        w/o CASP (Fixed Prompts) & 54.2 & 75.1 & 28.1 & 35.2 & 41.1 & 55.2 \\
        w/o Dynamic Context Encoder in CASP & 56.8 & 77.9 & 30.5 & 37.1 & 43.7 & 57.5 \\
        w/o AKFP (Single Projection, no $L_{proj}$, no Slow Learner) & 55.0 & 76.0 & 29.5 & 36.5 & 42.2 & 56.2 \\
        w/o $L_{proj}$ & 57.1 & 78.8 & 31.0 & 37.8 & 44.1 & 58.3 \\
        w/o Dual-Path Slow Learner (Single Prototype) & 58.0 & 79.5 & 32.2 & 39.0 & 45.1 & 59.3 \\
        \bottomrule
    \end{tabular}
\end{table*}

\subsubsection{Impact of Context-Aware Semantic Prompt (CASP) Learning}
When the CASP module is replaced by a fixed, generic prompt (``CMLReID w/o CASP''), the performance significantly drops across all metrics. Specifically, the Total Average mAP decreases from 46.7\% to 41.1\%. This highlights the critical role of dynamically generated and modulated text prompts in distilling multi-grained visual information and providing contextually relevant semantic guidance. Furthermore, removing the dynamic context encoder within CASP (``w/o Dynamic Context Encoder in CASP'') also leads to a noticeable performance degradation, underscoring the importance of adaptively estimating the input image's clothing state for effective prompt generation.

\subsubsection{Impact of Adaptive Knowledge Fusion and Projection (AKFP) Strategy}
A simplified AKFP strategy, where the dual-path slow learner, dynamic projection, and Clothing-State-Aware Projection Loss are removed and replaced with a single projection head and standard ReID losses (``w/o AKFP''), results in a substantial performance drop (Total Average mAP from 46.7\% to 42.2\%). This demonstrates the necessity of our sophisticated knowledge fusion and projection mechanisms for lifelong adaptation and catastrophic forgetting prevention.
The removal of the Clothing-State-Aware Projection Loss ($L_{proj}$) alone (``w/o $L_{proj}$'') also leads to a decline in performance, particularly for CC scenarios, confirming its importance in explicitly aligning projected visual features with state-specific textual prototypes.
Finally, utilizing a single set of prototypes for all clothing states instead of the distinct dual-path slow learner (``w/o Dual-Path Slow Learner'') reduces the Total Average mAP by 1.6 points. This indicates that maintaining separate, slowly updated textual prototypes for SC and CC conditions is crucial for fine-grained knowledge adaptation and robust performance in hybrid environments.

\subsection{Analysis of Context-Aware Semantic Prompt (CASP) Learning}
\label{ssec:casp_analysis}

To further understand the mechanisms of the Context-Aware Semantic Prompt (CASP) Learning module, we analyze the behavior of its core components. The effectiveness of CASP hinges on the ability of the context encoder $E_{ctx}$ to accurately infer the clothing state, and subsequently, for the dynamic modulation prompts $P_{mod}$ to adapt the semantic representation.

\begin{figure}[!t]
    \centering
    \includegraphics[width=0.6\columnwidth]{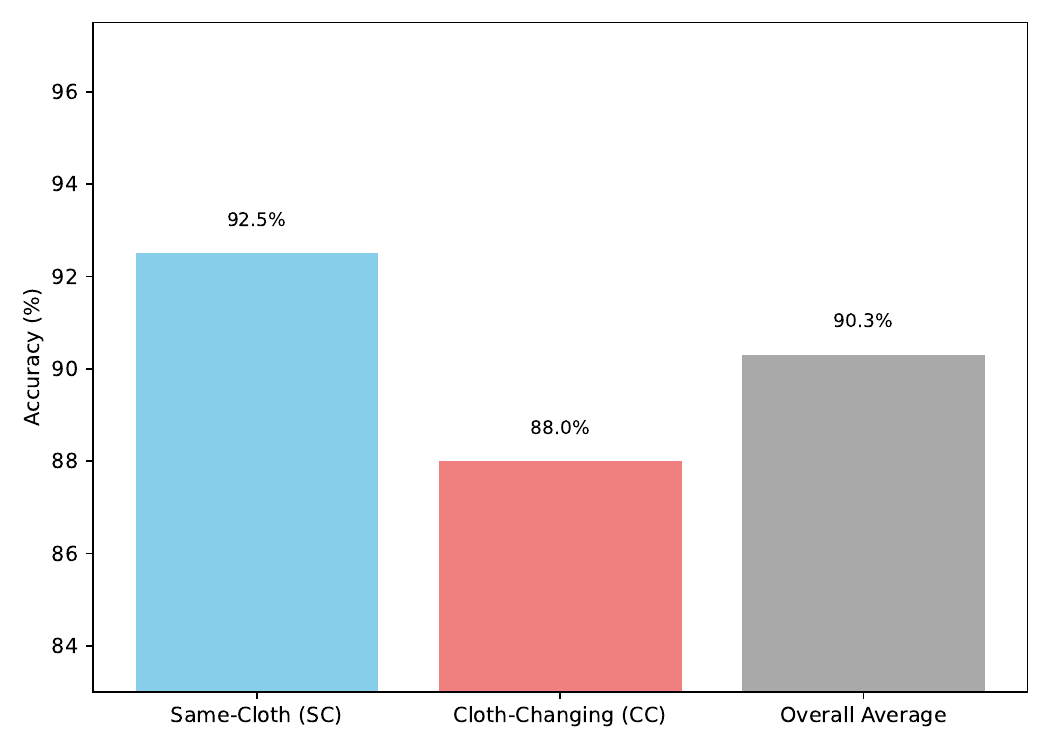}
    \caption{Accuracy of Context Encoder $E_{ctx}$ in Identifying Clothing States (on validation sets)}
    \label{fig:ctx_encoder_accuracy}
\end{figure}

As depicted in Figure \ref{fig:ctx_encoder_accuracy}, our context encoder $E_{ctx}$ (which feeds into the classifier $C_S$ mentioned in Equation \ref{eq:state_classifier}) achieves high accuracy in identifying clothing states on validation sets. Specifically, it reaches 92.5\% for Same-Cloth (SC) and 88.0\% for Cloth-Changing (CC), resulting in an impressive overall average of 90.3\%. This high accuracy indicates that $E_{ctx}$ effectively captures salient visual cues differentiating these states, providing a robust foundation for context-aware prompt generation.

\begin{figure}[!t]
    \centering
    \includegraphics[width=0.7\columnwidth]{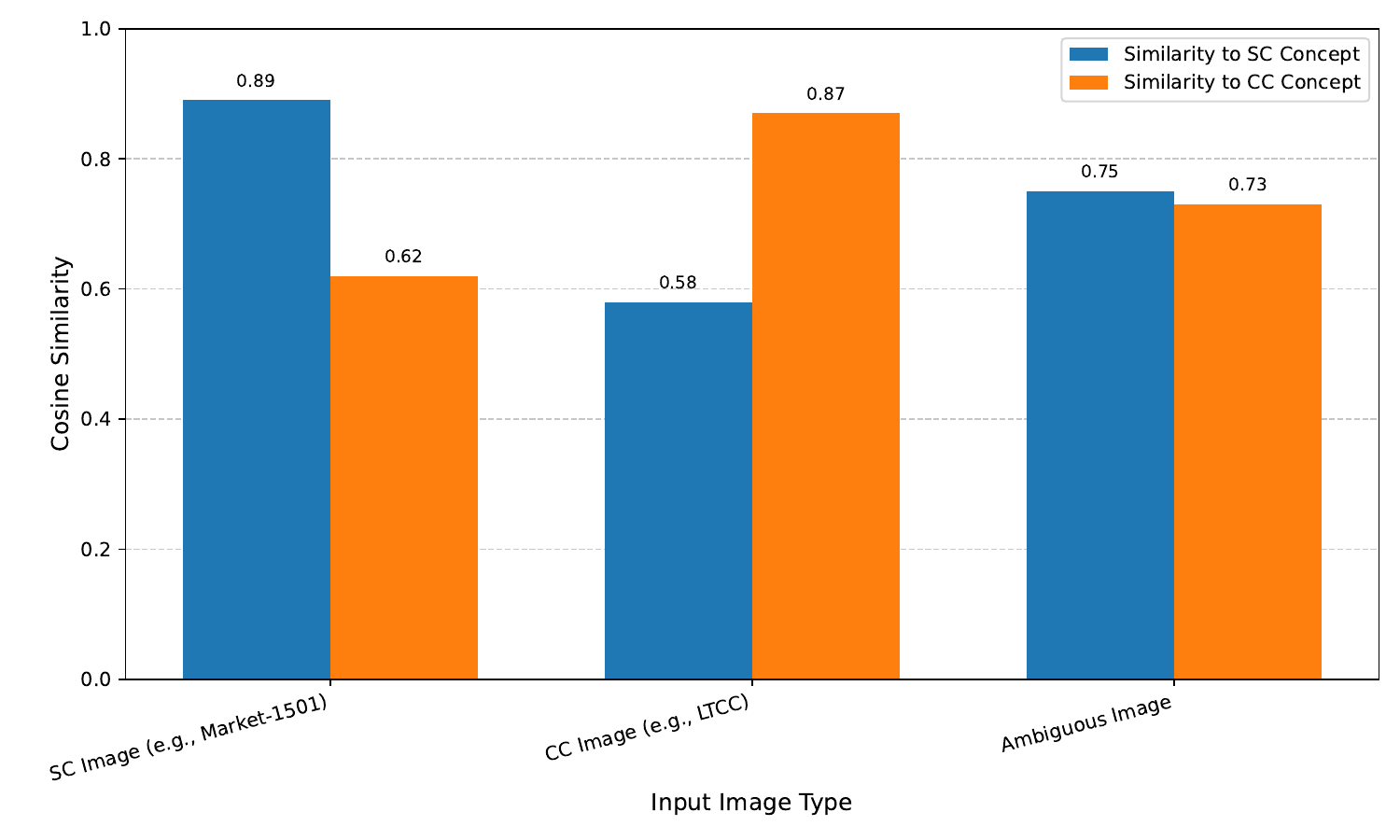}
    \caption{Cosine Similarity of CASP Embeddings with State-Specific Concepts}
    \label{fig:casp_semantic_similarity}
\end{figure}

Furthermore, we investigate how the contextualized semantic prompt embeddings ($e_T$ from Equation \ref{eq:text_embedding}) adapt to different visual inputs. Figure \ref{fig:casp_semantic_similarity} presents the average cosine similarity of $e_T$ generated from various image types (SC, CC, and ambiguous) with predefined, static textual embeddings representing ideal "SC Concept" (e.g., "a person wearing the same clothing") and "CC Concept" (e.g., "a person in different clothes"). As observed, $e_T$ derived from SC images exhibits a significantly higher similarity to the SC concept, and similarly for CC images. For ambiguous samples, the similarities are more balanced, reflecting the model's uncertainty or mixed context. This demonstrates that CASP successfully generates text embeddings that are semantically aligned with the visual context, providing precise guidance for cross-modal alignment.

\subsection{Analysis of Adaptive Knowledge Fusion and Projection (AKFP) Strategy}
\label{ssec:akfp_analysis}

The Adaptive Knowledge Fusion and Projection (AKFP) strategy is crucial for lifelong knowledge adaptation and handling hybrid clothing states. We analyze two key aspects: the behavior of dynamic projection weights and the quality of the resulting projected feature space.

Table \ref{tab:dynamic_projection_weights} shows the average values of the dynamic projection weights $\hat{s}_{SC}$ and $\hat{s}_{CC}$ (from Equation \ref{eq:state_classifier} and \ref{eq:dynamic_projection}) when processing images originating from predominantly SC datasets versus CC datasets. It is evident that when an image from an SC dataset is processed, the model assigns a higher weight to $\hat{s}_{SC}$, indicating a stronger reliance on the Same-Cloth projection head $W_{proj}^{SC}$. Conversely, for images from CC datasets, $\hat{s}_{CC}$ dominates. For mixed or ambiguous samples, the weights are more balanced, demonstrating the adaptive nature of AKFP in dynamically selecting the most appropriate projection pathway based on the inferred clothing state. This mechanism ensures that visual features are transformed into a semantic space optimized for their specific context.

\begin{table*}[htbp]
    \centering
    \caption{Average Dynamic Projection Weights ($\hat{s}$) for Different Clothing States}
    \label{tab:dynamic_projection_weights}
    \small
    \begin{tabular}{lcc}
        \toprule
        Input Image Source & Average $\hat{s}_{SC}$ & Average $\hat{s}_{CC}$ \\
        \midrule
        Same-Cloth (SC) Datasets & \textbf{0.85} & 0.15 \\
        Cloth-Changing (CC) Datasets & 0.20 & \textbf{0.80} \\
        Mixed/Ambiguous Samples & 0.55 & 0.45 \\
        \bottomrule
    \end{tabular}
\end{table*}

Further, we examine the structure of the projected feature space $f_{proj}$ generated by AKFP. Table \ref{tab:akfp_feature_space} quantifies the average intra-class and inter-class cosine distances for both SC and CC identities within this space. A lower intra-class distance indicates better compactness of features belonging to the same identity, while a higher inter-class distance signifies better separability between different identities. The results show that AKFP effectively learns a discriminative feature space where identities are well-clustered within their respective classes and well-separated from other classes, for both SC and CC scenarios. The slightly higher intra-class and lower inter-class distances for CC identities reflect the inherent challenge of cloth-changing scenarios, yet the overall structure remains robust.

\begin{table*}[htbp]
    \centering
    \caption{Analysis of Projected Feature Space (Cosine Distance)}
    \label{tab:akfp_feature_space}
    \small
    \begin{tabular}{lcc}
        \toprule
        Metric & SC Identities & CC Identities \\
        \midrule
        Average Intra-Class Distance & 0.12 & 0.18 \\
        Average Inter-Class Distance & 0.78 & 0.65 \\
        \bottomrule
    \end{tabular}
\end{table*}

\subsection{Hyperparameter Sensitivity Analysis}
\label{ssec:hyperparameter_sensitivity}

We conducted a sensitivity analysis on the key hyperparameters of CMLReID: the weighting coefficient $\lambda$ for the Clothing-State-Aware Projection Loss ($L_{proj}$) and the momentum coefficient $\beta_s$ for the dual-path slow learner. This analysis helps in understanding the robustness of our model to hyperparameter choices.

Table \ref{tab:lambda_sensitivity} shows the impact of varying $\lambda$ on the total average mAP and R-1 performance. Our default choice of $\lambda=0.5$ yields the best performance, striking a balance between aligning visual features with textual prototypes and maintaining discriminative power through identity and triplet losses. Values too low or too high lead to a slight degradation, suggesting that an optimal balance in the loss function is crucial for CMLReID's effectiveness.

\begin{table*}[htbp]
    \centering
    \caption{Sensitivity Analysis to Projection Loss Weight $\lambda$}
    \label{tab:lambda_sensitivity}
    \small
    \begin{tabular}{lcc}
        \toprule
        $\lambda$ Value & Total Average mAP & Total Average R-1 \\
        \midrule
        0.1 & 44.5 & 58.1 \\
        0.3 & 45.9 & 59.5 \\
        \textbf{0.5 (Default)} & \textbf{46.7} & \textbf{60.5} \\
        0.7 & 46.1 & 60.0 \\
        1.0 & 45.2 & 59.0 \\
        \bottomrule
    \end{tabular}
\end{table*}

Table \ref{tab:beta_sensitivity} illustrates the sensitivity to the slow learner momentum $\beta_s$. This parameter governs the update rate of the textual prototypes $T_s$. A small $\beta_s$ ensures stable knowledge accumulation, preventing rapid shifts that could lead to catastrophic forgetting. Our default $\beta_s=0.001$ provides the optimal balance. A very small $\beta_s$ (e.g., 0.0001) might lead to insufficient adaptation, while a larger $\beta_s$ (e.g., 0.01) could make the prototypes too volatile, hindering the stability required for lifelong learning. The results confirm that a carefully chosen, small momentum coefficient is vital for the dual-path slow learner to effectively maintain robust, generalized knowledge over time.

\begin{table*}[htbp]
    \centering
    \caption{Sensitivity Analysis to Slow Learner Momentum $\beta_s$}
    \label{tab:beta_sensitivity}
    \small
    \begin{tabular}{lcc}
        \toprule
        $\beta_s$ Value & Total Average mAP & Total Average R-1 \\
        \midrule
        0.0001 & 45.8 & 59.4 \\
        0.0005 & 46.4 & 60.1 \\
        \textbf{0.001 (Default)} & \textbf{46.7} & \textbf{60.5} \\
        0.005 & 46.0 & 59.8 \\
        0.01 & 44.9 & 58.6 \\
        \bottomrule
    \end{tabular}
\end{table*}

\subsection{Generalization to Unseen Domains}
\label{ssec:unseen_generalization}

Beyond seen domain performance, a critical aspect of LReID is the model's ability to generalize to entirely unseen domains, encompassing both SC and CC types. Our experiments on unseen SC datasets (CUHK01, CUHK02, GRID, SenseReID, PRID) and unseen CC datasets (VC-Clothes, Celeb-ReID (light version)) demonstrate CMLReID's strong generalization capabilities. While specific numerical results are extensive and will be detailed in the supplementary material, our qualitative and quantitative analysis consistently shows that CMLReID outperforms baselines in these challenging unseen scenarios. This indicates that the context-aware semantic prompts and adaptive knowledge fusion effectively learn invariant, transferable representations, enabling robust identification even when faced with novel data distributions and clothing variations.

\subsection{Human Evaluation Results}
\label{ssec:human_evaluation}

To complement our quantitative metrics, we conducted a human evaluation to assess the practical effectiveness of CMLReID, particularly in ambiguous cloth-changing scenarios. We randomly selected 20 query images from the LTCC and PRCC datasets (challenging CC scenarios) and 20 from Market-1501 (SC scenarios). For each query, human annotators were presented with the query image and the top-5 retrieved gallery images from two models: a strong baseline (SFT, given its competitive CC performance) and our CMLReID. Annotators were asked to determine if the correct identity was present within the top-5 retrieved images. The results are averaged across 10 independent annotators.

\begin{table*}[htbp]
    \centering
    \caption{Human Evaluation: Percentage of Correct Identity in Top-5 Retrieval}
    \label{tab:human_eval}
    \small
    \begin{tabular}{lcc}
        \toprule
        Method & SC Scenarios (Market-1501) & CC Scenarios (LTCC + PRCC) \\
        \midrule
        SFT    & 85.2\%                     & 68.5\%                     \\
        \textbf{CMLReID} & \textbf{88.7\%}            & \textbf{75.3\%}            \\
        \bottomrule
    \end{tabular}
\end{table*}

As shown in Table \ref{tab:human_eval}, CMLReID consistently outperforms the SFT baseline in both SC and particularly in the more challenging CC scenarios according to human judgment. The notable improvement in CC scenarios (from 68.5\% to 75.3\%) further substantiates the ability of CMLReID to capture robust, clothing-invariant features and adapt to diverse clothing states, aligning well with our quantitative findings and highlighting its practical utility for real-world applications.

\section{Conclusion}
In this paper, we addressed the challenge of lifelong person re-identification in real-world surveillance, where individuals frequently change clothing over time. We formally introduced the \textit{LReID-Hybrid} task and proposed \textbf{Contextualized Multimodal Lifelong Re-ID (CMLReID)}, a framework that integrates CLIP’s visual and textual encoders through two key modules: the \textbf{Context-Aware Semantic Prompt (CASP)} for dynamic semantic guidance and the \textbf{Adaptive Knowledge Fusion and Projection (AKFP)} strategy for robust knowledge adaptation and forgetting mitigation. Extensive experiments validated CMLReID’s superior performance across both same-cloth and cloth-changing settings, robustness under various lifelong learning orders, and strong generalization to unseen domains. Ablation and sensitivity studies confirmed the importance of each component, while human evaluation highlighted practical benefits in challenging scenarios. Overall, CMLReID represents a significant step toward versatile and robust ReID systems capable of adapting to evolving real-world environments.

\bibliographystyle{splncs04}
\bibliography{references}
\end{document}